\documentclass[letterpaper, 10 pt, conference]{ieeeconf}  

\IEEEoverridecommandlockouts                              

\usepackage{comment}
\usepackage{siunitx}
\usepackage{relsize}
\usepackage{ifthen}
\usepackage[colorinlistoftodos]{todonotes}
\usepackage[vlined,ruled,linesnumbered]{algorithm2e}
\usepackage{graphics} 
\usepackage{rotating}
\usepackage{color}
\usepackage{enumerate}
\usepackage[T1]{fontenc}
\usepackage{psfrag}
\usepackage{epsfig} 
\usepackage{booktabs}
\usepackage{graphicx}
\usepackage{url}
\usepackage{multirow}
\usepackage{array}
\usepackage{latexsym}
\usepackage{amsfonts}
\usepackage{amsmath}
\usepackage{amssymb}
\usepackage{xfrac}
\usepackage{xstring}
\usepackage{algorithmic}
\usepackage{multirow}
\usepackage{xcolor}
\usepackage{prettyref}
\usepackage{flexisym}
\usepackage{bigdelim}
\usepackage{breqn} 
\usepackage{listings}
\let\labelindent\relax
\usepackage{xspace}
\usepackage{bm}
\usepackage{soul}
\graphicspath{{./figures/}}
\usepackage{tikz}
\usetikzlibrary{matrix,calc}
\usepackage{lipsum}
\usepackage{mdwlist}

\let\itemize\undefined
\makecompactlist{itemize}{stditemize}
\usepackage{enumitem}

\usepackage[caption=false,font=footnotesize]{subfig}
\usepackage{graphicx}
\usepackage{epstopdf}

\usepackage{mathtools}
\usepackage{gensymb}

\makeatletter
\let\NAT@parse\undefined
\makeatother
\usepackage{hyperref}
\usepackage{cleveref}

\usepackage{censor}

\usepackage{cuted}
\usepackage{capt-of}

\usepackage{wrapfig}

\newcommand{\bdmath}{\begin{dmath}}
\newcommand{\edmath}{\end{dmath}}
\newcommand{\beq}{\begin{equation}}
\newcommand{\eeq}{\end{equation}}
\newcommand{\bdm}{\begin{displaymath}}
\newcommand{\edm}{\end{displaymath}}
\newcommand{\bea}{\begin{eqnarray}}
\newcommand{\eea}{\end{eqnarray}}
\newcommand{\beal}{\beq \begin{array}{lll}}
\newcommand{\eeal}{\end{array} \eeq}
\newcommand{\beas}{\begin{eqnarray*}}
\newcommand{\eeas}{\end{eqnarray*}}
\newcommand{\ba}{\begin{array}}
\newcommand{\ea}{\end{array}}
\newcommand{\bit}{\begin{itemize}}
\newcommand{\eit}{\end{itemize}}
\newcommand{\ben}{\begin{enumerate}}
\newcommand{\een}{\end{enumerate}}

\definecolor{myblue}{RGB}{65 105 225}

\newcommand{\hide}[1]{}

\newcommand{\hiddenText}{{\color{gray} hidden text.}}
\newcommand{\hideWithText}[1]{\hiddenText}

\title{\LARGE \bf
Kinodynamic Motion Retargeting
for Humanoid Locomotion via \\
Multi-Contact Whole-Body Trajectory Optimization
}

\author{Xiaoyu Zhang, Steven Haener, Varun Madabushi and Maegan Tucker
\thanks{Authors are with Georgia Institute of Technology, Atlanta, GA, USA}
\thanks{\{\texttt{\footnotesize xzhang636,shaener3,vmadabushi3,mtucker}\}@gatech.edu}}

\begin{document}

\maketitle
\thispagestyle{empty}
\pagestyle{empty}


\begin{abstract}
We present the Kinodynamic Motion Retargeting (KDMR) framework, a novel approach for humanoid locomotion that models the retargeting process as a multi-contact, whole-body trajectory optimization problem.
Conventional kinematics-based retargeting methods only look at kinematic feasibility, inevitably introducing physically inconsistent artifacts, such as foot sliding and ground penetration, that degrade the performance of downstream imitation learning policies.
To address these inconsistencies, KDMR extends beyond pure kinematics to explicitly enforce rigid-body dynamics and scheduled contact constraints. This requires explicit contact sequences, which are estimated systematically using human ground-reaction force data. The result is dynamically feasible motions both in terms of the rigid-body dynamics and the enforced contacts. The approach also enables an operator to add additional constraints or costs such as minimizing acceleration or enforcing torque limits, improving the overall quality of the reference motion.

We evaluate KDMR against the state-of-the-art baseline, GMR, across three key dimensions: 1) the dynamic feasibility and smoothness of the retargeted motions, 2) the accuracy of GRF tracking compared to raw source data, and 3) the training efficiency and final performance of downstream control policies trained via the BeyondMimic framework.
Experimental results demonstrate that KDMR significantly outperforms purely kinematic methods, yielding dynamically viable reference trajectories that accelerate policy convergence and enhance overall locomotion stability.
Our end-to-end pipeline will be open-sourced upon publication.
\end{abstract}

\begin{figure}
    \centering
    \includegraphics[width=0.82\linewidth]{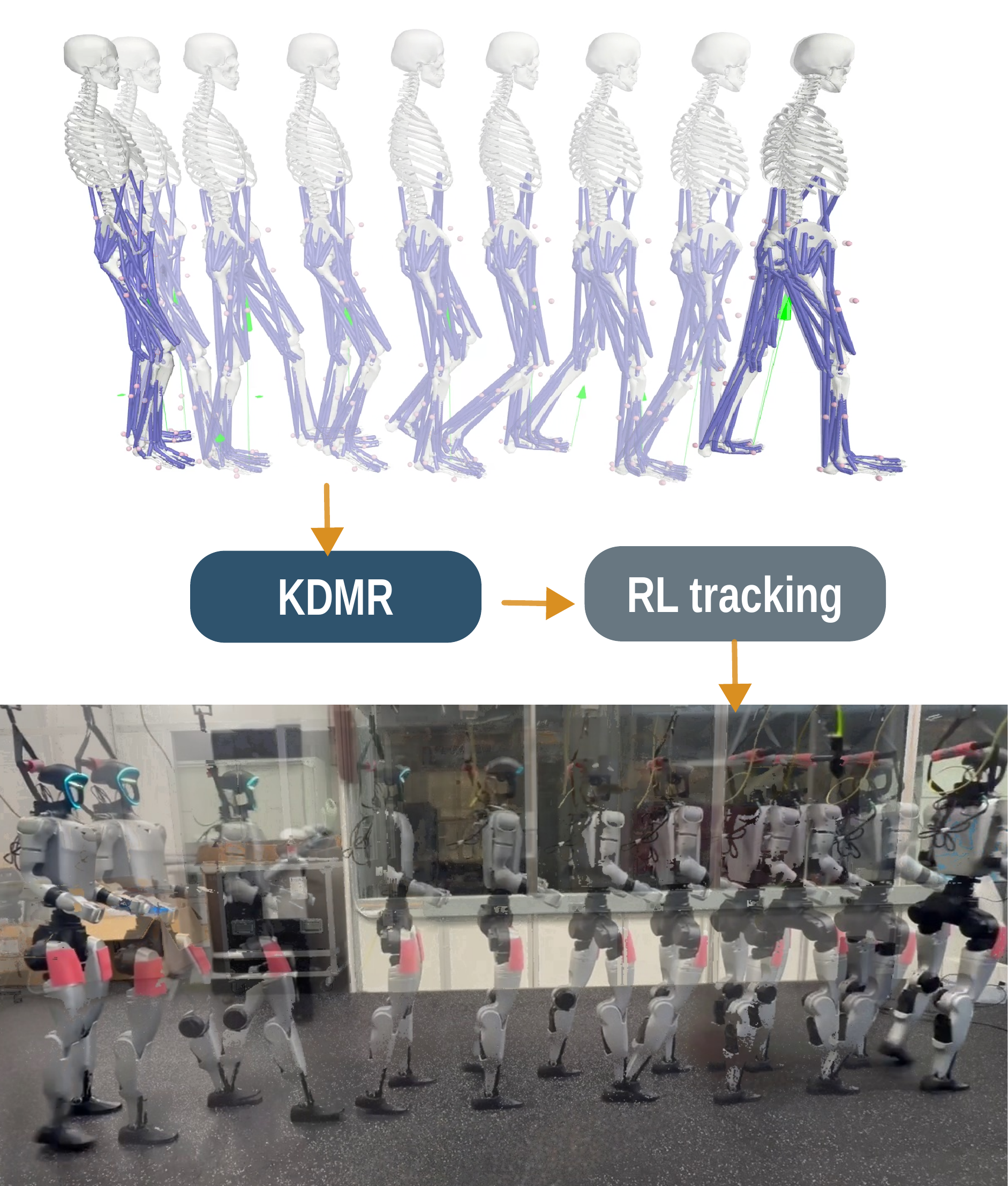}
    \caption{Given a source human motion and measured ground-reaction forces (GRFs), the proposed KDMR performs contact-aware trajectory optimization to generate a contact-consistent and dynamically feasible robot reference. A downstream RL policy tracks this reference on the real humanoid robot. 
    }
    \label{fig:title_image}
\end{figure}

\section{Introduction}\label{sec:intro}

Recent advances in humanoid locomotion have been largely driven by imitation learning and related approaches that transfer human-based motions, obtained primarily using motion capture (MoCap), to real-world humanoid robots.
By incorporating human motion references into reward design, humanoid robots can reproduce agile and natural human-like behaviors.
Furthermore, prior work has demonstrated that high-quality motion references can be leveraged to distilling reusable motion policies for generalizable control across a variety of downstream tasks \cite{peng2021amp,liao2025beyondmimic}.

However, human-based motions cannot be applied to robot motion directly; bridging the gap between human and robot morphology necessitates a motion retargeting process.
This process must rigorously account for discrepancies in joint limits, mass distribution, and actuation capabilities.
State-of-the-art retargeting methods only perform this mapping using kinematics, consequently introducing physical artifacts, such as foot sliding, ground penetration, and dynamically infeasible transitions \cite{araujo2025retargeting}.
If these artifacts are not explicitly resolved, they persist throughout the training process, forcing the learning agent to compensate for physically impossible references.
This inherently limits the sample efficiency and ultimate performance of the downstream control policies. 


Purely kinematics-based methods struggle to eliminate these artifacts because they neither enforce system dynamics nor capture the underlying contact forces. Moreover, simplified MoCap skeletons often omit articulated toes and detailed plantar contact geometry, thereby losing important foot--ground interaction information. GRF measurements provide complementary information about weight bearing and load transfer, which is essential for identifying heel--toe contact patterns in multi-contact humanoid motion~\cite{uchida2021biomechanics}.

To address these complexities, we propose a novel method for kinodynamic motion retargeting that explicitly certifies dynamic feasibility and holonomic constraint satisfaction for any arbitrary motion, including multi-contact locomotion.
Our approach formulates the retargeting procedure, leveraging human ground reaction forces as a primary physical anchor, as a whole-body trajectory optimization problem. This approach explicitly accounts for dynamic feasibility, enforces proper holonomic constraints for any contact sequence, and allows for additional constraints such as actuator limits. 
The main contributions of this work are as follows:

\begin{enumerate}
    \item A comprehensive pipeline, named Kinodynamic Motion Retargeting (KDMR), that jointly optimizes kinematics and whole-body dynamics. The result is smooth, dynamically feasible reference trajectories that bridge the morphological gap between humans and humanoid robots. 

    \item Demonstration that dynamics-consistent references improve sample efficiency and tracking performance across two motions: forward walking and twisting.

\end{enumerate}


\begin{figure*}[tb]
\vspace{2mm}
    \centering
    \includegraphics[width=0.99\linewidth]{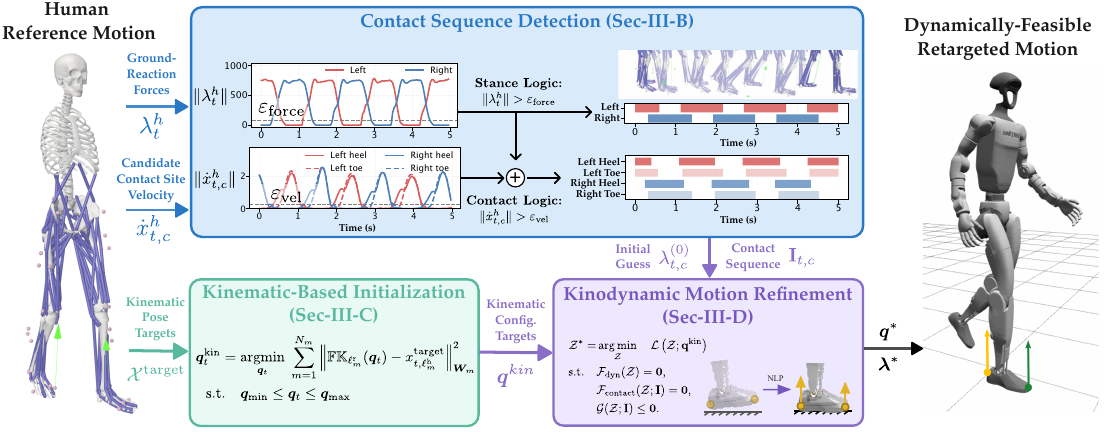}
    \caption{
Overview of KDMR. Human pose trajectories and GRF measurements are used to
infer the contact schedule and initialize the robot motion. The
initial trajectory is then refined through contact- and dynamics-constrained
whole-body optimization to produce a dynamically feasible robot reference.
    }
    \label{fig:overview}
    \vspace{-3mm}
\end{figure*}

\section{Related Work}\label{sec:lit_review}

\subsection{Imitation Learning for Humanoid Locomotion}\label{subsec:l4trakcing}
Imitation Learning (IL) is a common approach for producing natural, smooth motion tracking policies by imitating human data.
DeepMimic \cite{peng2018deepmimic} introduced a reinforcement learning (RL) approach to learn to reproduce motion references for different characters in a simulated environment using a goal-oriented reward function. Adversarial Motion Prior (AMP) \cite{peng2021amp} employs an adversarial discriminator to model motion style as a distributional prior, enabling better generalization beyond individual reference trajectories.

Building upon these prior efforts, several works \cite{RoboImitationPeng20, li2023learning, vollenweider2022advanced} demonstrated quadrupedal locomotion by leveraging animal motion capture (MoCap) data.
More recently, to enable human-like motion generation, methods such as ASAP \cite{he2025asap}, KungFuBot \cite{xie2025kungfubot}, Hub \cite{zhang2025hub}, OmniH2O \cite{he2024omnih2o}, UniTracker \cite{yin2025unitracker}, TWIST \cite{ze2025twist}, and BeyondMimic \cite{liao2025beyondmimic} have adopted imitation learning frameworks to achieve humanoid locomotion skills.
Although these approaches demonstrate agile and natural movements using retargeted kinematic trajectories, handling dynamic inconsistencies and retargeting artifacts during training remains challenging, critically impacts sample efficiency, learning stability, and performance \cite{araujo2025retargeting}.
In this paper, we mitigate retargeting artifacts and improve dynamic feasibility by casting the problem as a whole-body trajectory optimization problem.

\subsection{Kinematic Motion Retargeting}\label{subsec:kmr}
Motion retargeting originally aimed to transfer human motions to character animations in computer graphics \cite{popovic1999physically, tak2005physically}.
When extending this process to humanoid robots, additional challenges emerge due to morphological differences and dynamic inconsistencies, such as variations in degrees of freedom (DoFs), link lengths, inertial properties, and actuator limits.
Consequently, directly mapping joint values from human data to humanoid robots is insufficient \cite{cheng2024expressive, fu2024humanplus}, often leading to non-physical artifacts such as base floating, foot penetration, foot sliding, and non-smooth motions.

Recent works such as PHC \cite{Luo2023PerpetualHC} and its subsequent extensions \cite{he2024omnih2o, he2025asap, he2024hover}, aim to mitigate these morphological challenges. 
They formulate retargeting as an inverse kinematics (IK) problem solved via gradient descent through forward kinematics.
However, the resulting motions may be dynamically infeasible and still exhibit noticeable artifacts.
GMR \cite{araujo2025retargeting} improves upon this paradigm by incorporating a non-uniform local scaling strategy combined with a two-stage IK formulation to improve geometric consistency.
While they perform well in common human skeleton models, differences in skeleton structure  (e.g., additional intermediate foot joints) can introduce systematic height and grounding offsets, since ground alignment is often computed from specific joint positions rather than true contact geometry.
Despite these improvements, GMR still suffers from similar limitations, as ensuring dynamic feasibility and eliminating artifacts across the entire motion reference trajectories remain unresolved.

\subsection{Dynamics-Informed Motion Retargeting}

The artifacts arising from purely kinematic motion retargeting are closely coupled with violations of dynamic feasibility \cite{grizzle2014models,reher2020algorithmic}, as system dynamics and contact complementarity constraints are typically not enforced. Consequently, incorporating system dynamics into the retargeting process can substantially improve the physical plausibility of the generated motions and facilitate their transfer to real robots beyond visually convincing animations.

Several works have attempted to address this issue by refining kinematic reference trajectories through dynamics-aware optimization. For example, \cite{yoon2025spatio} enables robust transfer of complex animal motions to quadruped robots by jointly aligning spatial kinematics and temporal dynamics. 
Similarly, SPARK~\cite{wang2026spark} also refines kinematically retargeted motions through full-order kinodynamic trajectory optimization. It progressively solves kinematic trajectory optimization, inverse dynamics, and full kinodynamic optimization, assuming that foot-level contact labels are provided with the source motion. In contrast, our work focuses on transferring fine-grained foot--ground interaction by inferring a point-level heel--toe contact schedule from synchronized GRF and contact-point velocities. Therefore,
SPARK and KDMR address different aspects of dynamics-aware retargeting: progressive optimization and skeleton calibration in SPARK, and GRF-informed multi-point contact transfer in KDMR.

Other work uses ground-reaction-force measurement as an alternative source of contact information \cite{scherpereel2023human,han2023groundlink,ren2025motionpro,le2025physics}; KDMR leverages human GRF to systematically detect contact phase and as an optional force-initialization signal.
KDMR also computes an explicit, reproducible per-point contact schedule from measured marker velocity and models each foot's heel and toe as independently-active contact candidates, merging them into a single line-contact wrench only when both are planted. This enables changing contact modes, unlike SPARK's single rigid-foot contact wrench.

A complementary line of work enforces physical consistency through simulator rollouts. DynaRetarget \cite{dhedin2026dynaretarget} refines kinematic references using sampling-based trajectory optimization in MuJoCo \cite{todorov2012mujoco}, whereas Direct Dynamic Retargeting \cite{roux2026direct} directly optimizes task-space tracking through sampling-based MPC. Both resolve contacts implicitly through simulation and depend on sampling objectives and initialization. NMR \cite{zhao2026nmr} distills trajectories generated by simulated RL experts into a neural retargeter, enabling fast inference at the cost of morphology-specific training and implicit, distribution-dependent physical feasibility. In contrast, KDMR requires no learned retargeter and explicitly optimizes robot dynamics, contact forces, and scheduled contact constraints.

\section{Method}\label{sec:method}
Our method, KDMR, summarized in Fig.~\ref{fig:overview}, consists of two main components: contact force decomposition and kinodynamic motion retargeting. We first decompose the foot-level ground reaction force into contact-point forces and infer the corresponding contact sequence, as described in Sec.~\ref{subsec:contact-decomp}. Given this contact sequence and an initial kinematic retargeting trajectory described in Sec.~\ref{subsec:Kinematic}, we then solve a multi-contact trajectory optimization problem to generate the final dynamically feasible motion reference, as described in Sec.~\ref{subsec:dyna}.

    \vspace{-2mm}

\subsection{Problem Statement}
\label{subsec:problem-statement}

The input human biomechanics data consist of synchronized motion and ground reaction force (GRF) measurements over $T+1$ frames, indexed by $t=0,\ldots,T$. The motion is reconstructed from three-dimensional marker trajectories and fitted in OpenSim \cite{delp2007opensim} to a human skeletal model with $N_h$ degrees of freedom, yielding the generalized coordinates $\bm q_t^{\mathrm h}\in\mathbb R^{N_h}$ and velocities $\dot{\bm q}_t^{\mathrm h}$. The fitted motion is represented by a collection of human body poses $\mathcal X^{\mathrm{target}}=\{x^{\mathrm{target}}_{t,\ell^{\mathrm h}}\in SE(3)\mid \ell^{\mathrm h}\in\mathcal L^{\mathrm h},\ t=0,\ldots,T\}$, where $\mathcal L^{\mathrm h}$ is the set of selected human bodies and $x^{\mathrm{target}}_{t,\ell^{\mathrm h}}$ denotes the world-frame pose of human body $\ell^{\mathrm h}$ at frame $t$. The synchronized GRF measurements provide the net force acting on each foot, denoted by $\bm\lambda^{\mathrm h}_t=\{\bm\lambda^{\mathrm h}_{t,L},\bm\lambda^{\mathrm h}_{t,R}\}$.

The humanoid robot generally has a different kinematic structure, mass distribution, and foot geometry from the human subject. We model the robot as a floating-base system with configuration $\bm q_t=(\bm q_{t,b},\bm q_{t,l})\in\mathcal Q=SE(3)\times\mathcal Q^l$, where $\bm q_{t,b}\in SE(3)$ denotes the pelvis-attached floating-base pose and $\bm q_{t,l}\in\mathcal Q^l\subseteq\mathbb{R}^{N_l}$ contains the $N_l$ actuated joint coordinates. The corresponding generalized velocity and acceleration are denoted by $\dot{\bm q}_t$ and $\ddot{\bm q}_t$, respectively.


To transfer the human motion to the robot, we define a set of
corresponding rigid-body links
$\mathcal M=\{(\ell_m^{\mathrm h},\ell_m^{\mathrm r})\}_{m=1}^{N_{\mathcal M}}$,
where $\ell_m^{\mathrm h}$ and $\ell_m^{\mathrm r}$ denote the corresponding
human and robot links, respectively.
The human-link poses $\mathcal X^{\mathrm{target}}_{t,m}$ serve as task-space tracking targets
for the corresponding robot links.

Given $\mathcal X^{\mathrm{target}}$, $\bm\lambda^{\mathrm h}$, $\mathcal M$, and the robot model, our objective is to generate a dynamically feasible robot reference trajectory. Let $\mathcal Z=\{\bm q_t,\dot{\bm q}_t,\ddot{\bm q}_t,\bm\tau_t,\bm\lambda_t\}_{t=0}^{T}$ collect the robot configurations, velocities, accelerations, actuator torques, and contact forces. The retargeting problem is formulated abstractly as
\begin{equation}
\begin{aligned}
    \mathcal Z^{*}
    =
    \underset{\mathcal Z}{\operatorname{argmin}}
    \quad & 
    \mathcal L_{\mathrm{track}}
    \left(
        \mathcal Z;
        \mathcal X^{\mathrm{target}},
        \mathcal M
    \right)
    +
    \mathcal L_{\mathrm{reg}}(\mathcal Z)
    \\
    \mathrm{s.t.}\quad &
    \mathcal F_{\mathrm{dyn}}(\mathcal Z)=0,
    \\
    &
    \mathcal F_{\mathrm{contact}}(\mathcal Z;\bm I)=0,
    \\
    &
    \mathcal G(\mathcal Z;\bm I)\leq 0,
\end{aligned}
\label{eq:kdmr-general-problem}
\end{equation}
where $\bm I$ denotes the contact schedule constructed from the human motion and GRF measurements. The optimized trajectory should preserve the target human motion while satisfying the robot dynamics, scheduled contact constraints, and physical limits. The construction of $\bm I$ and the initialization of $\bm\lambda_t$ are described in Sec.~\ref{subsec:contact-decomp}, followed by the detailed trajectory optimization formulation.

\subsection{Contact Sequence Detection}
\label{subsec:contact-decomp}

The GRF measurements introduced in
Sec.~\ref{subsec:problem-statement} represent the resultant force acting on
each entire human foot, whereas the robot dynamics model applies forces at
four discrete contact points. We therefore infer a point-level binary contact
schedule over
$\mathcal C=\{\mathrm{LH},\mathrm{LT},\mathrm{RH},\mathrm{RT}\}$,
where $c\in\mathcal C$ indexes an individual contact point. Let
$\mathcal F=\{\mathrm L,\mathrm R\}$ denote the set of feet, with
$\mathcal C_{\mathrm L}=\{\mathrm{LH},\mathrm{LT}\}$ and
$\mathcal C_{\mathrm R}=\{\mathrm{RH},\mathrm{RT}\}$ denoting the contact
points associated with each foot.


We first determine the foot-level stance state $s_{t,c_f}\in\{0,1\}$ for $c_f\in\{L,R\}$ by thresholding the measured vertical GRF, such that $s_{t,c_f}=1$ when $\lambda^{\mathrm h,z}_{t,c_f}>\epsilon_{\mathrm{force}}$. This GRF gate prevents false-positive contacts when an airborne foot has a small instantaneous velocity, for example near the apex of its swing trajectory.

Within each detected stance phase, we use the velocities of the corresponding human heel and toe sites to determine which contacts participate. A rigid point in non-slipping contact with static ground should have approximately zero world-frame velocity. Motivated by this property and contact-candidate velocity representations such as CoCo-InEKF~\cite{baumgartner2026coco}, we compute the velocity of contact $c\in\mathcal C$ directly from the fitted OpenSim human model:
\begin{equation}
   \bm \dot{x}^{\mathrm h}_{t,c}
    =
    \bm J^{\mathrm h}_c
    \left(
        \bm q^{\mathrm h}_t
    \right)
    \dot{\bm q}^{\mathrm h}_t,
    \label{eq:contact_velocity}
\end{equation}
where $\bm J^{\mathrm h}_c$ is the world-frame translational Jacobian of human heel or toe site $c$. The binary contact state is defined as
\begin{equation}
    \mathbb I_{t,c}
    =
    \begin{cases}
        1,
        &
        s_{t,c_f}=1
        \ \text{and}\
        \left\|
            \bm x^{\mathrm h}_{t,c}
        \right\|_2
        <
        \epsilon_v,
        \\[2mm]
        0,
        &
        \text{otherwise},
    \end{cases}
    \label{eq:contact_state}
\end{equation}
We use the full three-dimensional velocity norm and the same threshold for both cyclic and non-cyclic motions. The resulting contact schedule is $\bm I=\{\mathbb I_{t,c}\mid t=0,\ldots,T,\ c\in\mathcal C\}$. The current implementation uses binary GRF and velocity gates and does not employ foot-height cues, learned contact confidence, or contact covariance estimation.

The contact schedule determines which point-force variables may be active.
For each foot $f\in\mathcal F$, let
\begin{equation}
    \mathcal A_{t,f}
    =
    \left\{
        c\in\mathcal C_f
        \mid
        \mathbb I_{t,c}=1
    \right\}
\end{equation}
denote its active contact points at frame $t$.
The measured foot-level GRF
is mass-scaled and equally distributed among the active contact points to initialize the
nonlinear program (NLP):
\begin{equation}
    \bm\lambda^{(0)}_{t,c}
    =
    \begin{cases}
        \displaystyle
        \frac{m^{\mathrm r}}{m^{\mathrm h}}
        \frac{\bm\lambda^{\mathrm h}_{t,f}}
             {|\mathcal A_{t,f}|},
        & c\in\mathcal A_{t,f},
        \\[2mm]
        \bm 0,
        & c\in\mathcal C_f\setminus\mathcal A_{t,f}.
    \end{cases}
    \label{eq:force_init}
\end{equation}
This equal split provides only an initial guess; the final point forces are
independently optimized subject to the robot dynamics and contact constraints.

\subsection{Kinematics-Based Initialization} \label{subsec:Kinematic}

Our method is initialized using the IK-based
retargeting approach. This stage serves only as an initial guess and
tracking prior for the dynamics-based retargeting stage described in
Sec.~\ref{subsec:dyna}. Given the target poses
$\mathcal X^{\mathrm{target}}$ and body correspondences $\mathcal M$
introduced in Sec.~\ref{subsec:problem-statement}, we compute the
initial robot configuration at each frame $t$ as

\begin{equation}
    \begin{aligned}
        \bm q^{\mathrm{kin}}_t
        =\;&
        \underset{\bm q_t}{\operatorname{argmin}}\;
        \sum_{m=1}^{N_{\mathcal M}}
        \left\|
            \bm e_m
            \left(
                \bm q_t,
                \mathcal X^{\mathrm{target}}_{t,m}
            \right)
        \right\|_{\bm W_m}^{2}
        \\
        \text{s.t.}\quad&
        \bm q_{\min}^{\mathrm l}
        \leq
        \bm q_t^{\mathrm l}
        \leq
        \bm q_{\max}^{\mathrm l},
    \end{aligned}
    \label{eqn:kopti-1}
\end{equation}
where $\bm q_t^{\mathrm l}$ denotes the local joint configuration and
$\bm W_m\succeq 0$ weights the task-space components of correspondence
$m$. Let
$\mathbb{FK}_{\ell_m^{\mathrm r}}(\bm q_t)
=(\bm p_{t,m}^{\mathrm r},\bm R_{t,m}^{\mathrm r})$
and
$\mathcal X^{\mathrm{target}}_{t,m}
=(\bm p_{t,m}^{\mathrm{target}},\bm R_{t,m}^{\mathrm{target}})$.
The pose residual is defined as

\begin{equation}
    \bm e_m
    =
    \begin{bmatrix}
        \bm p_{t,m}^{\mathrm r}
        -
        \bm p_{t,m}^{\mathrm{target}}
        \\
        \operatorname{Log}
        \left(
            (\bm R_{t,m}^{\mathrm{target}})^\top
            \bm R_{t,m}^{\mathrm r}
        \right)^\vee
    \end{bmatrix}.
\end{equation}

The resulting sequence
$\{\bm q^{\mathrm{kin}}_t\}_{t=0}^{T}$ matches the
selected human-link poses under the robot morphology and joint limits.
Because the configurations are obtained without enforcing temporal
dynamics, contact consistency, friction constraints, or torque
limits, the resulting motion may exhibit foot sliding, ground
penetration, temporal discontinuities, or dynamically infeasible
transitions. The subsequent dynamics-based retargeting stage reduces
these artifacts by enforcing trajectory-level dynamics and contact
constraints, using $\{\bm q^{\mathrm{kin}}_t\}_{t=0}^{T}$ as both the
configuration initial guess and the task-space tracking prior.

\subsection{Kinodynamic Motion Refinement}\label{subsec:dyna}
The kinematic trajectory $\{\bm q^{\mathrm{kin}}_t\}_{t=0}^{T}$ obtained from Sec.~\ref{subsec:Kinematic} provides a morphology-consistent warm start, but it does not enforce rigid-body dynamics, scheduled contact constraints, friction constraints, or actuator limits. Therefore, it may contain non-physical artifacts such as foot slip, ground penetration, and high-acceleration motions. We use $\bm q^{\mathrm{kin}}_t$ as the tracking prior and initialization for the following dynamics-based retargeting problem.

\subsubsection{Dynamics}
The robot's rigid-body dynamics follow the Euler-Lagrange equations of motion \cite{featherstone2008rigid}, 
\begin{equation}
    \bm D(\bm q_t)\ddot{\bm q}_t
    +\bm H(\bm q_t,\dot{\bm q}_t)
    =\bm B\bm\tau_t
    +\sum_{c\in\mathcal A_t}
    \bm J^{\mathrm r}_c(\bm q_t)^{\mathsf T}\bm\lambda_{t,c},
    \label{eq:dynamics}
\end{equation}
where $\bm D\in\mathbb R^{(N_l+6)\times(N_l+6)}$ is the generalized inertia matrix, $\bm H$ collects the Coriolis, centrifugal, and gravity terms, and $\bm B$ maps the actuator torques $\bm\tau_t$ to the generalized coordinates, and $\mathcal A_t\subseteq\mathcal C$ is the active contact set from Sec.~\ref{subsec:contact-decomp}. Each active contact $c$ enters the dynamics through its Jacobian $\bm J_c(\bm q_t)$, which maps generalized velocity to the world-frame velocity components conjugate to the contact wrench $\bm\lambda_{t,c}$, and is applied as a generalized force via $\bm J_c^{\mathsf T}\bm\lambda_{t,c}$. Every wrench not in $\mathcal A_t$ is fixed to zero, $\bm\lambda_{t,c}=\bm 0$, rather than left unconstrained, so an inactive contact cannot inject a physically ungrounded force into the dynamics. The contact-force variables are warm-started from the mass-scaled, equally-split measured GRF of Sec.~\ref{subsec:contact-decomp} (Eq.~\eqref{eq:force_init}); this is an initialization only; $\bm\lambda_{t,c}$ is subsequently optimized jointly with $\bm q_t,\dot{\bm q}_t,\ddot{\bm q}_t,\bm\tau_t$ subject to the constraints below.

\subsubsection{Holonomic Constraints}
The dimension of $\bm\lambda_{t,c}$ and its associated no-slip constraint depend on which contact points are active on a given foot.
For each foot, if only one contact point is active, the active contact point is modeled as a line contact and assigned a 5D wrench constraint $\bm\lambda_{t,c}=[F_x,F_y,F_z,M_x,M_z]\in\mathbb R^5$.
When the foot has both heel and toe planted, we merge merge the heel and toe line contacts into a single 6D wrench at the foot's midpoint frame,
$\bm\lambda_{t,f}^{\mathrm{mid}}=[F_x,F_y,F_z,M_x,M_y,M_z]\in\mathbb R^6$, and fix the individual point wrenches to zero. 
In this way, we handle the sixteen possible combinations of contact modes without introducing redundant constraints into the dynamic trajectory optimization problem.

For every active contact, the corresponding point (or midpoint) linear velocity is constrained to zero:
\begin{equation}
    \bm J_c(\bm q_t)\dot{\bm q}_t = \bm 0, \qquad c\in\mathcal A_t.
    \label{eqn:kc_v}
\end{equation}
This velocity-level no-slip constraint, together with the trapezoidal integration of Sec.~\ref{subsec:nlp}, implicitly keeps the contact acceleration small whenever a contact remains active across consecutive frames. For an inactive contact, the force is constrained by $\bm\lambda_{t,c}=\bm 0$ for $c\notin\mathcal A_t$. 

Each active wrench satisfies a linearized friction pyramid on its linear components,
\begin{align}
    \lambda^z_{t,c}\geq 0, \quad
    |\lambda^x_{t,c}|\leq
    \frac{\mu}{\sqrt{2}}\lambda^z_{t,c},
    \quad
    |\lambda^y_{t,c}|&\leq
    \frac{\mu}{\sqrt{2}}\lambda^z_{t,c},
    \label{eqn:friction}
\end{align}
with $\mu$ being the friction coefficient and $\lambda^x_{t,c},\lambda^y_{t,c},\lambda^z_{t,c}$ being the forward, lateral, and vertical force components of $\bm\lambda_{t,c}$. Additionally, zero moment point (ZMP) constraints are imposed:
\begin{equation}
    |M^x_{t,c}|\leq w\,\lambda^z_{t,c}, \quad |M^y_{t,f}|\leq \ell\,\lambda^z_{t,f}
    \label{eqn:zmp}
\end{equation}
with the foot's physical half-width $w$ and half-length $\ell$. Note that the second ZMP constraint (on $M^y$ is only applied for 6D wrenches. The yaw moment $M^z$ is left unconstrained in both cases.


\subsubsection{Objective}\label{subsec:objective}
For $t=0,\ldots,T-1$, the stage cost tracks the kinematic reference, regularizes acceleration, torque, and contact-force magnitude and smoothness, and penalizes the height of each active contact:
\begin{align}
\mathcal L_t={}&
\underbrace{\left\|\bm p_{t,b}-\bm p^{\mathrm{kin}}_{t,b}\right\|^2_{\bm w_p}}_{\text{Base position tracking}}
+\underbrace{
w_o
\left\|
\operatorname{Log}
\left(
(\bm R_{t,b}^{\mathrm{kin}})^\top
\bm R_{t,b}
\right)^\vee
\right\|_2^2}_{\text{Base orientation tracking}}
\nonumber\\
&+\underbrace{\left\|\bm q_{t,l}-\bm q^{\mathrm{kin}}_{t,l}\right\|^2_{\bm w_j}}_{\text{Joint tracking}}
+\underbrace{\left\|\ddot{\bm q}_t\right\|^2_{w_a}}_{\text{Acceleration reg.}}
+\underbrace{\left\|\bm\tau_t\right\|^2_{\bm w_\tau}}_{\text{Torque regularization}}
\nonumber\\
&+\underbrace{\sum_{c\in\mathcal C}\left\|\bm\lambda_{t,c}\right\|^2_{\bm w_\lambda}}_{\text{Force regularization}}
+\underbrace{\sum_{c\in\mathcal C}\left\|\bm\lambda_{t+1,c}-\bm\lambda_{t,c}\right\|^2_{\bm w_{\Delta\lambda}}}_{\text{Force smoothing}}
\nonumber\\
&+\underbrace{\left\|\dot{\bm q}_{t+1}-\dot{\bm q}_t\right\|^2_{\bm w_v}}_{\text{Velocity smoothing}}
+\underbrace{\sum_{c\in\mathcal A_t}\left|p^z_c(\bm q_t)\right|^2_{w_h}}_{\text{Contact-height penalty}}.
\label{eqn:stage_cost}
\end{align}
Here $\bm p_t,\hat{\bm\xi}_t$ are the base position and orientation quaternion, and $\bm q_{t,l}$ the actuated joint coordinates.

\subsubsection{NLP Implementation}\label{subsec:nlp}

Configurations, velocities, and accelerations obey trapezoidal integration for $t=0,\ldots,T-1$:
\begin{align}
    \bm q_{t+1}
    &=\bm q_t+\frac{\Delta t}{2}
    \left(\dot{\bm q}_{t+1}+\dot{\bm q}_t\right),
    \label{eqn:integral1}\\
    \dot{\bm q}_{t+1}
    &=\dot{\bm q}_t+\frac{\Delta t}{2}
    \left(\ddot{\bm q}_{t+1}+\ddot{\bm q}_t\right).
    \label{eqn:integral2}
\end{align}

The dynamics-based retargeting problem is then the following NLP:
\begin{equation}
\begin{aligned}
    \mathcal Z^*=\;&
    \underset{\{\bm q_t,\dot{\bm q}_t,\ddot{\bm q}_t,
    \bm\tau_t,\bm\lambda_t\}}
    {\operatorname{argmin}}
    \quad \sum_{t=0}^{T-1}\mathcal L_t
    \\
    \text{s.t.}\quad
    &\text{dynamics in \eqref{eq:dynamics}},\\
    &\text{contact constraints in \eqref{eqn:kc_v}},\\
    &\text{force constraints in \eqref{eqn:friction}--\eqref{eqn:zmp}},\\
    &\text{integration constraints in \eqref{eqn:integral1}--\eqref{eqn:integral2}},\\
    &\|\bm\tau_t\|_\infty\leq\tau_{\max},\\
    &\bm q_{\min}\leq\bm q_t\leq\bm q_{\max}.
\end{aligned}
\label{eqn:cost}
\end{equation}
Here $\bm\lambda_t=\{\bm\lambda_{t,c}\}_{c\in\mathcal C}$ collects all four contact forces at frame $t$, and $\tau_{\max}$ and $(\bm q_{\min},\bm q_{\max})$ denote the actuator-torque and configuration limits, respectively.

\begin{figure*}[!t]
\centering
\subfloat[Base position and orientation\label{fig:robot_base_p}]{
  \includegraphics[width=0.43\textwidth]{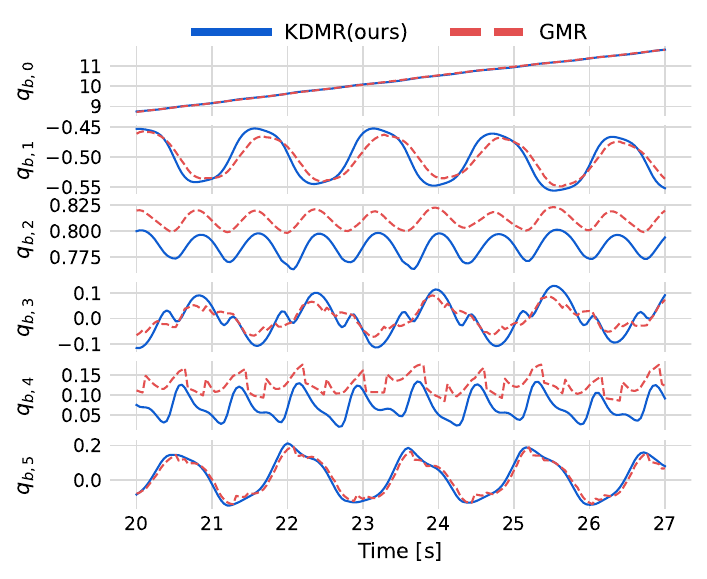}
}
\hspace{-3mm}
\vspace{-3mm}
\subfloat[Base linear and angular velocity\label{fig:robot_base_v}]{
  \includegraphics[width=0.43\textwidth]{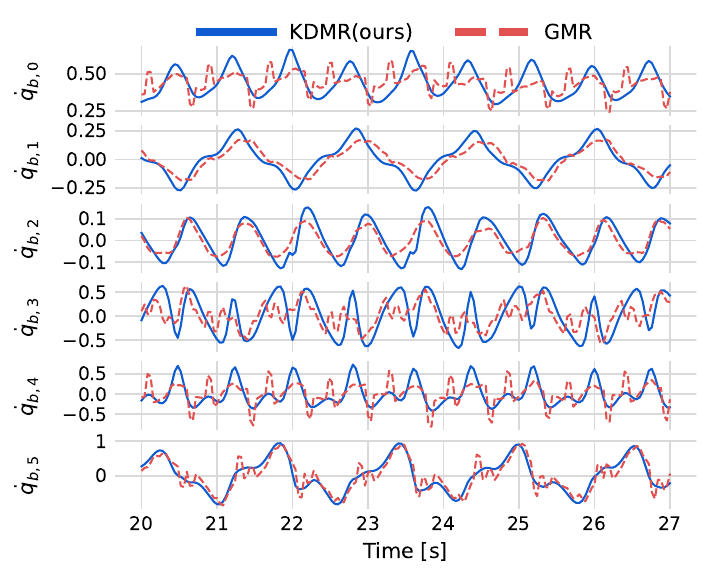}
}
\hspace{-3mm}
\subfloat[Joint angles for one leg\label{fig:robot_dof_p}]{
  \includegraphics[width=0.43\textwidth]{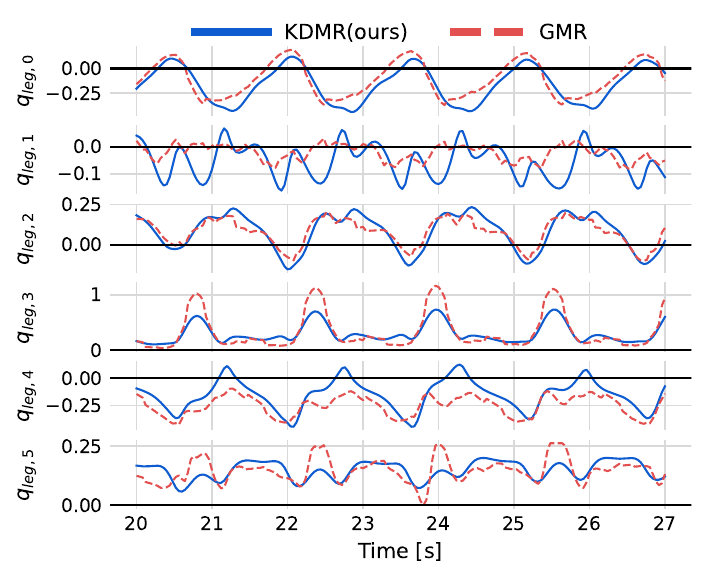}
}\hspace{-3mm}
\subfloat[Joint velocity for one leg\label{fig:robot_dof_v}]{
  \includegraphics[width=0.43\textwidth]{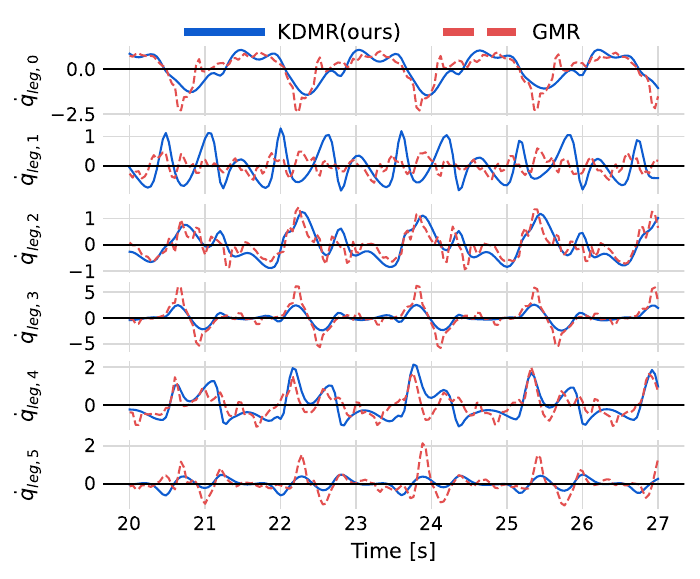}
}
\caption{Comparison of retargeted trajectories produced by KDMR and GMR over a time segment (20--27~s) of the full motion sequence. (a) Robot base configuration $\bm{q}_{b}$, consisting of three translational and three rotational coordinates. (b) Corresponding base velocity $\dot{\bm{q}}_{b}$. (c) Joint configuration $\bm{q}_{\mathrm{leg}}$ of one leg, ordered from the hip-pitch joint to the ankle-roll joint. (d) Corresponding joint velocity $\dot{\bm{q}}_{\mathrm{leg}}$. Here, $q_{b,i}$ and $\dot{q}_{b,i}$ denote the $i$-th base coordinate and its velocity, while $q_{\mathrm{leg},i}$ and $\dot{q}_{\mathrm{leg},i}$ denote the position and velocity of the $i$-th leg joint, respectively, for $i=0,\ldots,5$.}
\label{fig:base-plot}
\end{figure*}


\section{Results}
\label{sec:results}
We evaluate KDMR by retargeting walking and whole-body
twisting motions from an open-source biomechanics dataset~\cite{scherpereel2023human}
to the Unitree G1 humanoid.
We compare KDMR with GMR~\cite{araujo2025retargeting}, a kinematic retargeting
baseline.
We first describe the experimental setup in Sec.~\ref{exp:setup}.
We then evaluate the retargeted motion, contact behavior, and downstream policy
performance in Secs.~\ref{exp:dynamic_smoothness}--\ref{exp:policy}. Finally, we
present the real-robot deployment results in Sec.~\ref{exp:sim2real}

\subsection{Implementation Details} \label{exp:setup}
We implement our optimization framework using CasADi \cite{Andersson2019} to formulate and solve the nonlinear programming (NLP) problem, alongside Pinocchio \cite{carpentier2019pinocchio} for efficiently computing the rigid-body kinematics and dynamics.
This implementation makes introducing a new target robot straightforward.
Specific cost function weights and other implementation details can be found on our repository\footnote{\url{https://anonymous.4open.science/r/KDMR}}.
For the downstream evaluation, the motion tracking policies are trained using the default BeyondMimic pipeline \cite{liao2025beyondmimic}, implemented by the MJLab framework \cite{zakka2026mjlab}. To ensure a fair comparison, we carefully tuned GMR and applied the minimum-height correction described in the original paper to reduce floating and ground-penetration artifacts.



\begin{table*}[t]
    \centering
    \caption{
        Comparison of KDMR and GMR across mean errors: global body position error ($E_{g-bpe}$), body position error ($E_{bpe}$), and joint position error ($E_{jpe}$). Each error is computed over 100 rollouts with different seeds.
    }
    \label{tab:kdmr_gmr_comparison}

    \setlength{\tabcolsep}{4.5pt}
    \renewcommand{\arraystretch}{1.15}

    \begin{tabular}{
        l
        l
        *{4}{S[table-format=2.1]}
        *{4}{S[table-format=2.1]}
        *{4}{S[table-format=2.1]}
    }
        \toprule
        & &
        \multicolumn{4}{c}{$E_{\mathrm{g\text{-}bpe}}$ (\si{\milli\meter})}
        & \multicolumn{4}{c}{$E_{\mathrm{bpe}}$ (\si{\milli\meter})}
        & \multicolumn{4}{c}{$E_{\mathrm{jpe}}$ (\si{\milli\radian})} \\
        
        \cmidrule(lr){3-6}
        \cmidrule(lr){7-10}
        \cmidrule(lr){11-14}

        Motion
        & Method
        & {Mean} & {Median} & {Min.} & {Max.}
        & {Mean} & {Median} & {Min.} & {Max.}
        & {Mean} & {Median} & {Min.} & {Max.} \\
        \midrule

        \multirow{2}{*}{Walk}
        & GMR
        & {14.7} & {14.6} & {12.9} & {17.9}
        & {12.8} & {12.7} & {10.8} & {16.0}
        & {40.6} & {40.8} & {38.5} & {42.7} \\

        & KDMR
        & {\bfseries 9.1} & { \bfseries9.1} & {\bfseries 7.5} & {\bfseries 12.9}
        & {\bfseries 8.3} & {\bfseries 8.2} & {\bfseries 6.6} & {\bfseries 12.3}
        & {\bfseries 29.4} & {\bfseries 29.4} & {\bfseries 27.7} & {\bfseries 31.7} \\

        \midrule

        \multirow{2}{*}{Twister}
        & GMR
        & 22.35 & 21.88 & 19.25 & 27.32
        & 18.22 & 17.99 & 15.47 & 23.49
        & 46.31 & 46.21 & 43.31 & 51.62 \\
        & KDMR
        & {\bfseries 11.9} & {\bfseries 11.7} & {\bfseries 9.8}  & {\bfseries 14.04}
        & {\bfseries 10.8} & {\bfseries 10.5} & {\bfseries 8.4}  & {\bfseries 14.8}
        & {\bfseries 33.6} & {\bfseries 33.5} & {\bfseries 31.8} & {\bfseries 36.3} \\

        \bottomrule
    \end{tabular}
    \vspace{-4mm}
\end{table*}

\subsection{Dynamic Feasibility and Motion Smoothness}\label{exp:dynamic_smoothness}

We first evaluate whether KDMR improves the dynamic feasibility and temporal smoothness of the retargeted motion while preserving its principal kinematic characteristics. 
We use the treadmill walking sequence for this comparison because its repeated gait cycles make contact violations and trajectory discontinuities particularly visible.

As shown in Fig.~\ref{fig:base-plot}, KDMR produces the same overall motion pattern as GMR, while introducing localized corrections to the base and lower-limb trajectories. The largest differences occur around the ankle joints, whose configurations directly affect the heel and toe positions. These deviations are necessary for satisfying the robot dynamics and holonomic constraints for the prescribed contact schedule.

\begin{figure}
    \centering
    \includegraphics[width=0.82\linewidth]{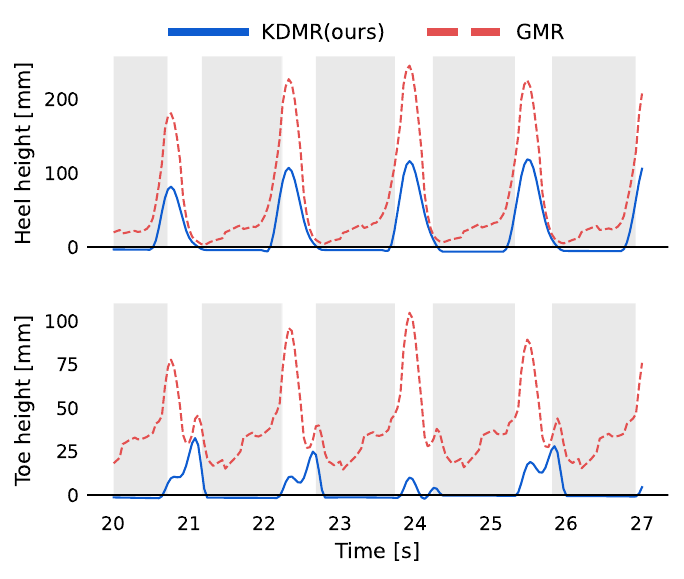}
    \caption{The height of heel and toe with contact window.
    }
    \label{fig:foot_height}
\end{figure}

The foot-height comparison in Fig.~\ref{fig:foot_height} further illustrates this distinction. During each stance phase, KDMR constrains the active heel or toe contact points to remain on the ground surface. In contrast, GMR does not explicitly impose stance-phase contact constraints and therefore frequently produces foot-floating artifacts. Moreover, even when a GMR contact point is geometrically close to the ground, physical contact is not guaranteed because its contact force and consistency with the whole-body dynamics are not enforced. KDMR instead jointly enforces the active contact heights, feasible contact forces, and whole-body equations of motion. The foot-height result therefore provides geometric evidence of the improved contact feasibility, while the dynamics and contact-force constraints establish its dynamic consistency.

The velocity profiles in Figs.~\ref{fig:robot_base_v} and~\ref{fig:robot_dof_v} show a corresponding improvement in motion smoothness. GMR exhibits abrupt variations and high-magnitude velocity spikes, particularly in the lower-limb joints around contact transitions. KDMR substantially reduces these fluctuations and produces more regular, temporally continuous velocity profiles.

Overall, these results show that KDMR preserves the gait style while producing smoother trajectories and more physically consistent stance contacts than the kinematic baseline.

\subsection{Downstream Policy Tracking Performance}
\label{exp:policy}

\begin{figure}
    \centering
    \includegraphics[width=0.82\linewidth]{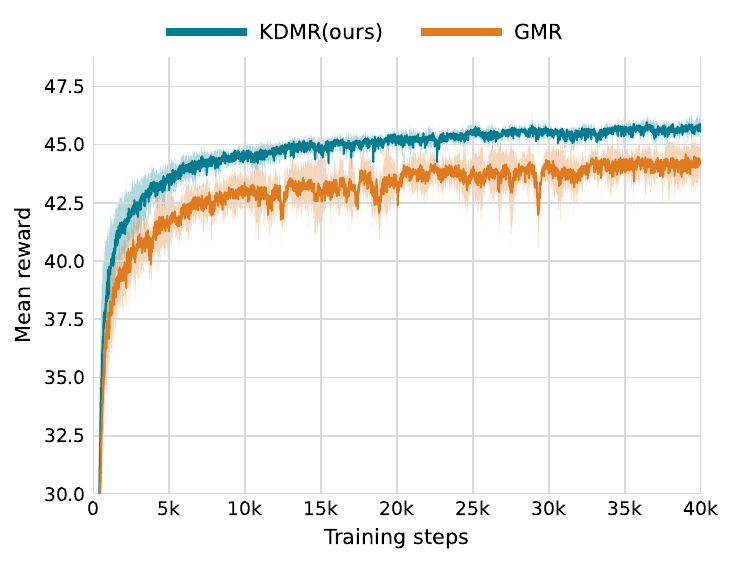}
    \caption{Mean episode rewards over three independent training runs. Solid lines show the mean, and shaded regions show one standard deviation.}
    \label{fig:mean_reward}
\end{figure}

Next, we evaluate whether the improved contact and dynamic consistency of KDMR benefits downstream policy learning.
For each motion, we train separate policies to track the references generated by GMR and KDMR.
The two methods use identical observations, actions, reward terms, network architectures, PPO hyperparameters, training budgets, and simulation settings. The only difference is the retargeted reference motion used for training. Figure~\ref{fig:mean_reward} shows that policies trained with KDMR references achieve higher rewards with fewer environment samples and converge earlier than those trained with GMR references.

After training, each policy is evaluated against its corresponding
training reference over 100 rollouts with different random seeds.
Table~\ref{tab:kdmr_gmr_comparison} reports the body-position
and joint-position tracking errors. KDMR achieves lower errors across
all three metrics for both motions. For walking, KDMR reduces the mean
global body position error ($E_{\mathrm{g\text{-}bpe}}$), body position
error ($E_{\mathrm{bpe}}$), and joint position error
($E_{\mathrm{jpe}}$) by $38.1\%$, $35.2\%$, and $27.6\%$,
respectively. For the twisting motion, the reductions
are $46.8\%$, $40.7\%$, and $27.4\%$.

Together, the training curves and rollout results show that KDMR
references are learned with better sampling efficiency and
reproduced more accurately under the same policy-learning setup.
Combined with the contact-consistency and smoothness results in
Sec.~\ref{exp:dynamic_smoothness}, these findings suggest that removing
dynamically inconsistent artifacts, such as stance-phase foot floating
and abrupt trajectory variations, provides a more learnable reference
and reduces the compensatory behavior required from the tracking policy.

\subsection{Real-World Transfer}\label{exp:sim2real}
We deploy the policies $\pi_{\mathrm{GMR}}$ and $\pi_{\mathrm{KDMR}}$ for both motions on the real Unitree G1 robot, with the performance best shown in the supplemental video. Both policies can be transferred without real-world fine-tuning.
For the twisting motion, however, $\pi_{\mathrm{GMR}}$ exhibits
more fragile behavior and is more susceptible to falling during
abrupt whole-body transitions, whereas $\pi_{\mathrm{KDMR}}$
produces more stable executions. This qualitative observation
suggests that smoother and dynamically consistent references can
facilitate real-world transfer for motions involving rapid
whole-body transitions.


\section{Limitation}
\label{sec:limitation}

\begin{table}[t]
    \centering
    \caption{
Runtime comparison of GMR and KDMR for generating retargeted reference trajectories of \textit{800} frames
    }
    \label{tab:runtime}
    \vspace{-0.5em}
    \begin{tabular}{llcc}
        \toprule
        Method & Motion  & Runtime [s] \\
        \midrule
        \multirow{2}{*}{GMR}
            & Walking  & \textit{7} \\
            & Twist    & \textit{10} \\
        \midrule
        \multirow{2}{*}{KDMR}
            & Walking  & \textit{88.6} \\
            & Twist    & \textit{282.4} \\
        \bottomrule
    \end{tabular}
    \vspace{-0.5em}
\end{table}

A primary limitation of KDMR is its computational cost. As shown in Table~\ref{tab:runtime}, KDMR is slower than the purely kinematic GMR baseline because it solves a constrained trajectory optimization problem involving whole-body dynamics and contact forces. Nevertheless, KDMR is designed as an offline motion-retargeting method, and real-time computation is not its objective. Each reference trajectory is generated only once and can then be reused throughout policy training and deployment. As the training curves and rollout results show, this one-time preprocessing cost is a favorable trade-off for the improved sample efficiency and tracking quality it yields downstream.

We also note that kinodynamic retargeting may not be necessary for every type of motion.
We expect its benefits to be most pronounced for motions with complex contact sequences, where enforcing dynamic feasibility improves the smoothness and success rate of the downstream tracking policy.

\newpage
\section{Conclusion}
\label{sec:conclusion}
In this work, we introduce a kinodynamic motion retargeting pipeline for humanoid whole-body locomotion tasks, formulated through a multi-contact dynamic model. Our experiments demonstrate that the proposed method significantly outperforms purely kinematic baselines in mitigating physical artifacts arising from dynamic inconsistencies. Extending beyond generating visually plausible kinematics, our framework successfully recovers accurate foot contact events and the corresponding ground reaction force magnitudes directly from human source data. 


\bibliographystyle{IEEEtran}
\bibliography{references}

\end{document}